\pdfoutput=1

\documentclass[11pt]{article}

\usepackage[final]{acl}

\usepackage{times}
\usepackage{latexsym}

\usepackage[T1]{fontenc}

\usepackage[utf8]{inputenc}

\usepackage{microtype}

\usepackage{inconsolata}

\usepackage{graphicx}

\usepackage{booktabs}
\usepackage{multirow}
\usepackage{enumitem}
\usepackage{comment}

\usepackage{amsmath,amssymb,graphicx}
\usepackage{lipsum}
\usepackage{lipsum}
\usepackage{algorithm}
\usepackage{algpseudocode}

\usepackage[normalem]{ulem}
\useunder{\uline}{\ul}{}

\usepackage{placeins}
\usepackage{subcaption}

\title{Model Merging for Knowledge Editing}

\author{
  \textbf{Zichuan Fu\textsuperscript{1}\thanks{Work was conducted during the internship at Tencent Jarvis Lab.}},
  \textbf{Xian Wu\textsuperscript{2}\footnotemark[2]},
  \textbf{Guojing Li\textsuperscript{1}},
  \textbf{Yingying Zhang\textsuperscript{2}},
  \textbf{Yefeng Zheng\textsuperscript{2,3}},\\
  \textbf{Tianshi Ming\textsuperscript{4}},
  \textbf{Yejing Wang\textsuperscript{1}},
  \textbf{Wanyu Wang\textsuperscript{1}},
  \textbf{Xiangyu Zhao\textsuperscript{1}\thanks{Corresponding authors.}}
\\
\\
  \textsuperscript{1} City University of Hong Kong
  \textsuperscript{2} Tencent Jarvis Lab  \\
  \textsuperscript{3} Westlake University 
  \textsuperscript{4} Tongji University 
\\
  \small{
  \texttt{
    \href{mailto:zc.fu@my.cityu.edu.hk}{zc.fu@my.cityu.edu.hk},
    \href{mailto:kevinxwu@tencent.com}{kevinxwu@tencent.com},
    \href{mailto:xianzhao@cityu.edu.hk}{xianzhao@cityu.edu.hk}
  }}
}

\begin{document}
\maketitle
\begin{abstract}
Large Language Models (LLMs) require continuous updates to maintain accurate and current knowledge as the world evolves. While existing knowledge editing approaches offer various solutions for knowledge updating, they often struggle with sequential editing scenarios and harm the general capabilities of the model, thereby significantly hampering their practical applicability.
This paper proposes a two-stage framework combining robust supervised fine-tuning (R-SFT) with model merging for knowledge editing. Our method first fine-tunes the LLM to internalize new knowledge fully, then merges the fine-tuned model with the original foundation model to preserve newly acquired knowledge and general capabilities. 
Experimental results demonstrate that our approach significantly outperforms existing methods in sequential editing while better preserving the original performance of the model, all without requiring any architectural changes. Code is available at \href{https://github.com/Applied-Machine-Learning-Lab/MM4KE}{Applied-Machine-Learning-Lab/MM4KE}.

\end{abstract}

\section{Introduction}
\label{introduction}

\begin{figure*}[t]
    \centering
    \includegraphics[width=\linewidth]{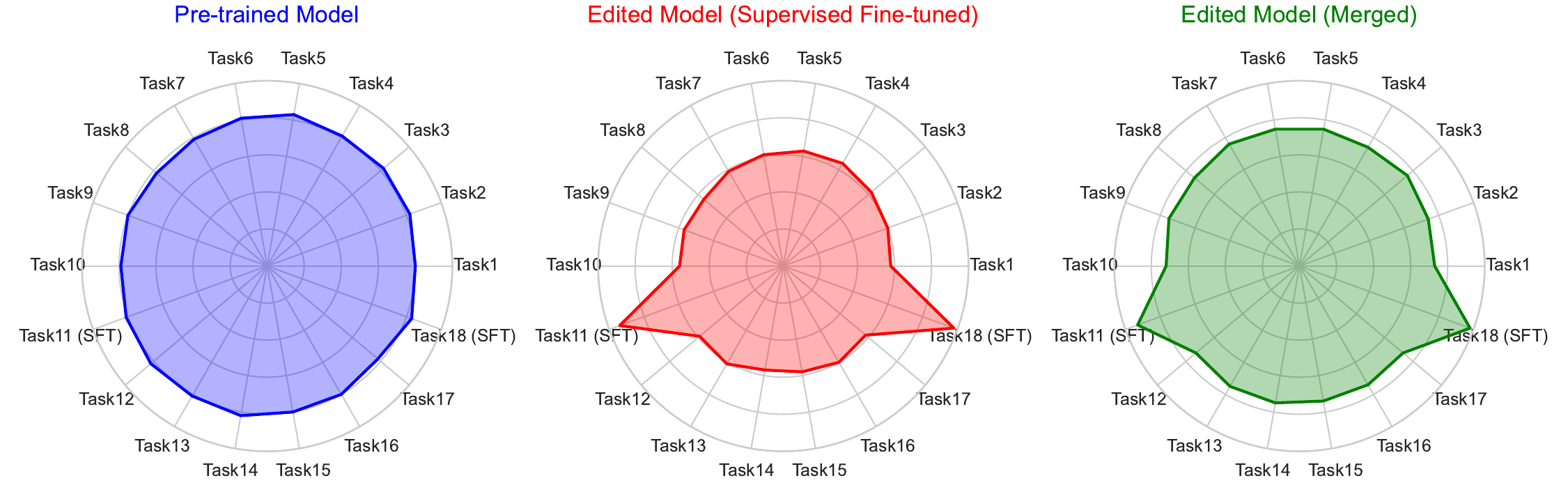}
    \caption{The illustration of three radar charts demonstrates the performance distribution across multiple tasks. The left chart shows the pre-trained model excelling in general tasks but limited in specific tasks (SFT). The middle chart represents the fine-tuned model with enhanced specific task performance at the cost of general capabilities. The right chart illustrates the merged model that successfully maintains both general and specific task performance.}
    \label{fig:demo}
\end{figure*}

Large Language Models (LLMs) have revolutionized Natural Language Processing (NLP) by capturing vast amounts of world knowledge and exhibiting impressive generalization capabilities~\cite{llmsurvey,uni-ctr,graph}. Recent advancements in both architecture design and training strategies have enabled LLMs such as GPT-4~\cite{gpt4} to engage in human-like dialogue and solve complex real-world problems. 

However, when deployed in dynamic real-world environments, LLMs often face challenges of maintaining current and accurate knowledge~\cite{model-editing-survey}. For example, models can quickly become outdated regarding political developments, technological innovations, or evolving natural disasters; they may also retain inaccurate historical details or harmful content that needs timely removal to ensure safe and reliable outputs.

To tackle these challenges, knowledge editing has emerged as an effective solution for efficiently updating or correcting specific information in pre-trained language models. These approaches can be broadly categorized into three main categories~\cite{knowedit}. Memory-based methods primarily rely on fine-tuning mechanisms to store and update knowledge in the model's parameters~\cite{GRACE}. Meta-learning approaches leverage auxiliary networks to learn how to generate precise weight updates for knowledge editing. Locate-then-edit methods directly identify and modify specific components within the model architecture to update factual associations. Each of these approaches offers distinct strategies for modifying model behavior.

However, these existing approaches still face several significant limitations. 
First, most editing methods exhibit poor performance in sequential editing and often suffer from weak generalization capabilities. 
As a result, they struggle to effectively inject large amounts of knowledge into the models, limiting their practical applicability~\cite{wang2024wise,InstructEdit}.
Second, after knowledge editing, models often experience degradation in their general capabilities, as the editing process typically focuses only on targeted knowledge without considering its impact on unrelated knowledge~\cite{ROME,MEMIT}.

To address the above limitations, we propose a simple yet effective knowledge editing framework integrating Robust Supervised Fine-Tuning (R-SFT) with Model Merging techniques. Specifically, we employ R-SFT, a fine-tuning strategy that selectively optimizes only the Feed-Forward Networks (FFNs) in a single transformer layer. We use iterative sample-wise optimization paired with an early-stopping mechanism to avoid overfitting. Subsequently, we merge the fine-tuned model with the original foundation model through scaling and sparsity-driven pruning, recovering general capabilities compromised during fine-tuning while effectively retaining acquired factual edits. Extensive experimental evaluations demonstrate significant performance improvements over existing methods across sequential editing tasks, superior preservation of general capabilities, and no architectural modifications are required.

\begin{itemize}[leftmargin=*]
    \item We propose R-SFT, an efficient fine-tuning approach leveraging sample-wise iterative optimization with early stopping to ensure precise and efficient knowledge acquisition.
    \item We apply model merging to mitigate the negative impact of fine-tuning on the general capabilities of LLMs, providing a simple but effective solution without any architectural modifications.
    \item Experimental results show that our method outperforms existing approaches in sequential editing while maintaining the general capabilities.
\end{itemize}

\section{Methodology}
\label{sec:method}

This section introduces the proposed two-stage framework for knowledge editing, which includes R-SFT and model merging. 

\subsection{Robust Supervised Fine-tuning}
\label{ssec:sft}

Existing knowledge editing methods face significant challenges in sequential edits, often requiring complex architectural modifications that limit their practical applicability. Therefore, in the first stage of our framework, we propose Robust Supervised Fine-tuning (R-SFT), a robust knowledge learning fine-tuning paradigm designed to overcome these limitations while maintaining simplicity and effectiveness, as detailed in Algorithm~\ref{alg:rsft}.

\begin{algorithm}[t]
\caption{Procedure of Robust Supervised Fine-Tuning (R-SFT)}
\label{alg:rsft}
\begin{algorithmic}[1]
\Require Foundation model $\theta_{\mathrm{base}}$, dataset $\mathcal{D} = \{ s_n \}_{n=1}^N$, learning rate $\eta$, early stop threshold $\tau$, max epochs $E$, max steps per sample $K$
\State Initialize model parameters: $\theta^{(0)} \leftarrow \theta_{\text{base}}$
\State Set global iteration counter: $t \leftarrow 0$
\For{$e = 1$ \textbf{to} $E$} \Comment{Iterate epochs}
    \For{$n = 1$ \textbf{to} $N$} \Comment{Iterate samples}
        \For{$k = 1$ \textbf{to} $K$} \Comment{Iterative steps}
            \State $\mathcal{L}_{n} = -\log P(\mathbf{a}_n|\mathbf{q}_n;\theta^{(t)})$
            \If {$\mathcal{L}_{n} <\tau$} \Comment{Early stopping}
                \State \textbf{break}
            \Else
                \State $\theta^{(t+1)} \leftarrow \theta^{(t)} - \eta \nabla_{\theta}\mathcal{L}_{n}$
                \State $t \leftarrow t + 1$
            \EndIf
        \EndFor
    \EndFor
\EndFor
\State \textbf{return} fine-tuned parameters $\theta_{\mathrm{sft}} \leftarrow \theta^{(t)}$
\end{algorithmic}
\end{algorithm}

Specifically, given a pre-trained foundation model $\theta_{\mathrm{base}}$ and an editing dataset $\mathcal{D} = \{(\mathbf{q}_n, \mathbf{a}_n)\}_{n=1}^N$, where each sample includes a question $\mathbf{q}_n$ and its corresponding targeted answer $\mathbf{a}_n$, R-SFT aims to update the model parameters to encode the provided factual information accurately. The objective follows the standard supervised fine-tuning (SFT), minimizing the negative log-likelihood of the correct output given the input:
\begin{equation}
    \mathcal{L}_n(\theta) = -\log P(\mathbf{a}_n|\mathbf{q}_n;\theta)
\end{equation}
For each sample, we iteratively update the parameters via gradient descent with learning rate $\eta$:
\begin{equation}
    \theta^{(t+1)} = \theta^{(t)} - \eta \nabla_{\theta}\mathcal{L}_n(\theta^{(t)})
\end{equation}
where $t$ is the global iteration counter.

The key difference between R-SFT and conventional SFT is the sample-level consecutive training with an early-stop mechanism. In each epoch, each sample is optimized consecutively for at most $K$ steps, stopping early if the loss decreases below the threshold $\tau$:
\begin{equation}
    k_n^{*} = \min\{k \mid \mathcal{L}_n(\theta^{(t+k)}) < \tau \text{ and } 1 \leq k \leq K \}
\end{equation}
where $k_n^{*}$ denotes the real number of gradient update steps performed on the $n$-th sample within the epoch. A sample that satisfies the early stop criterion remains available in subsequent epochs, allowing periodic validation to avoid forgetting.

Furthermore, based on insights from existing research~\cite{ROME}, we restrict R-SFT solely to the Feed-Forward Networks (FFN) of the fifth transformer layer, which has been proven to be optimal for editing performance and efficiency.

After completing the R-SFT process over $E$ epochs, we obtain a fine-tuned model $\theta_{\mathrm{sft}}$ that thoroughly and reliably captures the desired knowledge edits.
This fine-tuned model, along with the original pre-trained foundation model $\theta_{\mathrm{base}}$, forms the foundation for our subsequent merging stage.

\subsection{Model Merging}
\label{ssec:fusion}

In the second stage, the fine-tuned model is merged with the foundation model.
While R-SFT effectively teaches the model new knowledge, it typically comes at the cost of degrading the model's general capabilities.
Therefore, we employ model merging, including scaling and pruning, to restore these fundamental capabilities while preserving the newly acquired knowledge.

Our merging approach employs a weighted average of the original and fine-tuned models, essentially applying \textbf{scaling} to the fine-tuned model:
\begin{equation}
\theta_{\mathrm{edited}} = \alpha\theta_{\mathrm{base}} + (1-\alpha)\theta_{\mathrm{sft}}, \alpha \in (0,1)
\label{eq:merge}
\end{equation}
where a scaling parameter controls the preservation-editing trade-off.
This equation can be further reformulated to highlight the parameter difference:

\begin{equation}
\theta_{\mathrm{edited}} = \theta_{\mathrm{base}} + (1-\alpha)(\theta_{\mathrm{sft}} - \theta_{\mathrm{base}})
\label{eq:delta}
\end{equation}
where $\Delta\theta = \theta_{\mathrm{sft}} - \theta_{\mathrm{base}}$ represents the knowledge delta, the parameter changes that encode the new knowledge acquired during R-SFT.

To further reduce the interference of knowledge delta on general capabilities, we apply \textbf{pruning} to the knowledge delta:
\begin{equation}
    \theta_{\mathrm{edited}} = \theta_{\mathrm{base}} + (1-\alpha) \cdot \text{Top}_p(\theta_{\mathrm{sft}} - \theta_{\mathrm{base}})
    \label{eq:prune}
\end{equation}
The pruning operation keeps the top $p\%$ of parameters with the highest magnitude changes in each parameter matrix, while setting the rest to zero. 

This process induces a high degree of sparsity in the knowledge delta, ensuring that only the most impactful modifications are retained. Such sparsity not only reduces the risk of interference with the pretrained model’s general capabilities, but also suppresses noisy updates introduced by training samples or the fine-tuning process. 

Finally, the merged model can preserve general capabilities, while effectively incorporating the newly acquired knowledge from R-SFT.

\subsection{Industrial Application Prospect}
\label{ssec:emerging}

Real-world industry applications require specialized LLMs capable of performing domain-specific tasks without losing foundational general-purpose capabilities such as comprehension and logic reasoning. Foundation models typically lack domain-specific accuracy, while traditional fine-tuning methods introduce significant limitations: fine-tuning solely on vertical data often causes catastrophic forgetting~\cite{forget}, whereas hybrid training with extensive general and domain data incurs prohibitive computational costs.

The proposed R-SFT enables efficient domain-specific data optimization. Meanwhile, the model merging strategy combines the fine-tuned domain-specific models and the foundation model, thereby integrating specialized domain knowledge without sacrificing general linguistic reasoning capabilities. We have successfully delivered multiple specialized models tailored to distinct professional domains, demonstrating improved performance on their targeted tasks and maintaining the general language processing competencies necessary for practical industrial applications.

\section{Experiments}
\label{experiments}

\begin{table*}[htbp]
\centering
\small
\caption{Performance comparison of merging methods for sequential knowledge editing. The best values are highlighted in bold, while the second-best values are underlined. Column ``Base'' represents the foundation model.}
\label{tab:overall}
\resizebox{\textwidth}{!}{ 
\begin{tabular}{@{}lrcccccccc@{}}
\toprule
\textbf{DataSet} & \textbf{Metric} & \textbf{Base} & \textbf{KN} & \textbf{ROME} & \textbf{MEMIT} & \textbf{LoRA} & \textbf{SFT} & \textbf{R-SFT} & \textbf{Merged} \\ \midrule
\multicolumn{10}{c}{Edited Knowledge} \\ \midrule
\textbf{ZsRE} & Edit Succ. $\uparrow$ & - & 6.66 & 14.53 & 3.11 & 98.06 & {\ul 99.39} & \textbf{99.82} & 96.95 \\
 & Generalization $\uparrow$ & - & 6.79 & 12.53 & 3.09 & 73.52 & 85.13 & \textbf{93.29} & {\ul 91.58} \\
 & Portability $\uparrow$ & - & 10.43 & 2.32 & 1.06 & 20.90 & 24.40 & \textbf{47.48} & {\ul 39.63} \\
 & Locality $\uparrow$ & - & 7.54 & 1.13 & 1.20 & 5.28 & 12.65 & \textbf{36.69} & {\ul 26.42} \\
 & Fluency $\uparrow$ & - & 421.73 & \textbf{535.50} & {\ul 477.30} & 411.80 & 414.58 & 441.53 & 420.49 \\ \midrule
\multicolumn{10}{c}{General Capabilities} \\ \midrule
\textbf{C-Eval} & Accuracy $\uparrow$ & \textbf{79.57} & 25.78 & 24.59 & 25.11 & 70.43 & 31.43 & 78.97 & {\ul 79.35} \\
\textbf{CoQA} & EM $\uparrow$ & {\ul 56.82} & 24.42 & 0.00 & 0.00 & 53.98 & 0.63 & 51.80 & \textbf{62.10} \\
 & F1 $\uparrow$ & {\ul 72.60} & 34.13 & 0.07 & 0.00 & 69.10 & 1.39 & 63.57 & \textbf{75.18} \\
\textbf{DROP} & EM $\uparrow$ & 0.23 & 0.03 & 0.00 & 0.00 & \textbf{1.96} & 0.09 & 0.67 & {\ul 1.9} \\
 & F1 $\uparrow$ & 7.10 & 2.07 & 0.32 & 0.00 & \textbf{13.90} & 0.21 & 8.23 & {\ul 10.8} \\
\textbf{SQuAD 2.0} & EM $\uparrow$ & 10.02 & 0.33 & 1.02 & \textbf{43.80} & 11.03 & 5.15 & 8.20 & {\ul 17.82} \\
 & F1 $\uparrow$ & 21.15 & 3.15 & 1.08 & \textbf{43.80} & 22.45 & 5.39 & 12.90 & {\ul 25.02} \\
\textbf{LogiQA} & Accuracy $\uparrow$ & \textbf{37.94} & 21.51 & 20.28 & 22.12 & 31.03 & 24.12 & 24.42 & {\ul 33.03} \\ \bottomrule
\end{tabular}
}
\end{table*}

In this section, our experiments are structured around the following research questions (RQs):
\begin{itemize}[leftmargin=*]
    \item \textbf{RQ1:} How does our model merging approach perform on the ZsRE dataset compared to baseline methods, and how does it impact the model's general capabilities?
    \item \textbf{RQ2:} How effective is our model merging approach across other knowledge editing datasets in KnowEdit?
    \item \textbf{RQ3:} How hyperparameter settings for robust model fine-tuning affect the accuracy and generalization ability of knowledge editing.
    \item \textbf{RQ4:} How do different components of our framework individually contribute to the overall performance of the edited model?
\end{itemize}

\subsection{Experimental Settings}


\subsubsection{Datasets}

We select KnowEdit~\cite{knowedit} for knowledge editing tasks, mainly on ZsRE dataset~\cite{zsre}.
For general ability evaluation, we use C-Eval~\cite{ceval}, CoQA~\cite{coqa}, DROP~\cite{DROP}, SQuAD 2.0~\cite{squad} and LogiQA~\cite{LogiQA}.

\subsubsection{Baselines}

In our experiments, we compare our approach against two main categories of locate-then-edit methods: 1) classic knowledge editing methods (ROME~\cite{ROME}, MEMIT~\cite{MEMIT}) that directly modify model parameters associated with specific facts, and 2) fine-tuning approaches (LoRA~\cite{Lora}) that update knowledge through training.

\subsubsection{Implementation Details}
\label{ssec:implementation-details}

We conduct experiments using EasyEdit~\cite{EasyEdit} for evaluating various knowledge editing methods, and employ the lm-evaluation-harness\footnote{\href{https://github.com/EleutherAI/lm-evaluation-harness}{https://github.com/EleutherAI/lm-evaluation-harness}} for assessing general model capabilities. R-SFT is implemented through LLaMA Factory~\cite{LlamaFactory} and mergeKit~\cite{MergeKit} for training and merging respectively. We use Qwen2.5-7B-Instruct~\cite{qwen2} as our foundation model.

\subsubsection{Evaluation Metrics}

We evaluate the models using two sets of metrics. To evaluate editing performance, we use five metrics: Edit Success (Edit Succ. or Succ.), Generalization (Gen.), Portability (Port.), Locality (Loc.) and Fluency (Flu.). 
The detailed definitions are provided in Appendix~\ref{ssec:app-metrcs}.
To assess the preservation of general capabilities, we use Accuracy for classification tasks (C-Eval, LogiQA), and both Exact Match (EM) and F1 scores for question-answering benchmarks (CoQA, DROP, SQuAD 2.0).

\begin{table}[ht]
\footnotesize
\centering
\caption{Editing performance on additional KnowEdit datasets using our framework.}
\label{tab:rq2}
\resizebox{\linewidth}{!}{  
\begin{tabular}{@{}llccc@{}}
\toprule
\textbf{DataSet} & \textbf{Metric $\uparrow$} & \textbf{SFT} & \textbf{R-SFT} & \textbf{Merged} \\ \midrule
\multirow{4}{*}{\textbf{WikiData$_{recent}$}} & Edit Succ. & 79.46 & \textbf{99.97} & 96.62 \\
 & Portability & 46.59 & 58.26 & \textbf{62.95} \\
 & Locality & 28.50 & 31.87 & \textbf{41.62} \\
 & Fluency & 428.95 & 461.51 & \textbf{592.02} \\ \midrule
\multirow{3}{*}{\textbf{WikiBio}} & Edit Succ. & 66.06 & \textbf{99.48} & 96.54 \\
 & Locality & 40.16 & 64.30 & \textbf{75.18} \\
 & Fluency & 626.60 & \textbf{628.77} & 626.71 \\ \midrule
\multirow{4}{*}{\textbf{WikiData$_{counter}$}} & Edit Succ. & 50.67 & \textbf{99.06} & 84.02 \\
 & Portability & 34.56 & \textbf{60.61} & 51.98 \\
 & Locality & 15.75 & 26.36 & \textbf{41.98} \\
 & Fluency & 479.81 & 601.02 & \textbf{614.64} \\ \bottomrule
\end{tabular}
}
\end{table}

\begin{table}[ht]
\centering
\caption{Effect of different hyperparameter settings on the editing performance.}
\label{tab:combined}

\begin{subtable}{\linewidth}
    \centering
    \small
    \caption{Early stopping loss threshold.}
    \label{tab:early}
    \begin{tabular}{@{}lccccc@{}}
    \toprule
    \textbf{Threshold} & \textbf{Succ.} & \textbf{Gen.} & \textbf{Port.} & \textbf{Loc.} & \textbf{Flu.} \\ \midrule
    None & 68.90 & 65.76 & 24.40 & 12.65 & \textbf{514.58} \\
    0.01 & 75.74 & 73.28 & 39.86 & 27.84 & 435.20 \\
    0.02 & 78.06 & 74.87 & 41.77 & 26.14 & 437.26 \\
    0.05 & 79.61 & 76.22 & 42.53 & 33.00 & 420.41 \\
    0.1 & \textbf{80.07} & \textbf{76.76} & 44.33 & 32.18 & 400.84 \\
    0.2 & 78.87 & 75.04 & \textbf{46.14} & \textbf{34.76} & 411.97 \\ 
    \bottomrule
    \end{tabular}
\end{subtable}

\vspace{1em} 

\begin{subtable}{\linewidth}
    \centering
    \caption{Number of epochs and steps.}
    \label{tab:epoch-step}
    \resizebox{\linewidth}{!}{
    \begin{tabular}{@{}llccccc@{}}
    \toprule
    \textbf{Epochs} & \textbf{Steps} & \textbf{Succ.} & \textbf{Gen.} & \textbf{Port.} & \textbf{Loc.} & \textbf{Flu.} \\ \midrule
    1 & 30 & 75.74 & 73.28 & 39.86 & 27.84 & 435.20 \\
    2 & 15 & 93.89 & 89.94 & 40.96 & 26.33 & 422.18 \\
    3 & 10 & 96.95 & 91.58 & 39.63 & 26.42 & 420.49 \\
    5 & 6 & 99.42 & 93.56 & 41.81 & 25.84 & 439.81 \\
    10 & 3 & 99.82 & \textbf{93.56} & 43.50 & 30.48 & 417.75 \\
    30 & 1 & \textbf{99.84} & 93.30 & \textbf{46.87} & \textbf{33.81} & \textbf{509.18} \\ 
    \bottomrule
    \end{tabular}
    }
\end{subtable}
\end{table}

\subsection{Overall Performance (RQ1)}

As shown in Table~\ref{tab:overall}, our empirical evaluation reveals several important findings regarding knowledge editing performance and preservation of general capabilities across different methods.

For knowledge editing, R-SFT exhibits superior editing performance across primary metrics, with the merged model maintaining the second-highest performance in most editing dimensions. Regarding general capabilities, the merged model effectively retains the foundation model's general capabilities, demonstrating comparable performance on C-Eval and enhanced results on CoQA. This suggests our merging strategy successfully addresses the common trade-off between knowledge editing and general capability preservation.

Notably, MEMIT performs surprisingly well on SQuAD 2.0, and LoRA achieves strong results on DROP. This is largely because the foundation model originally performed poorly on these tasks, making it more sensitive to minor perturbations introduced during editing. These edits may alter the model’s answering behavior in a way that coincidentally improves the evaluation metrics, rather than reflecting true methodological superiority.

\subsection{Knowledge Editing Performance (RQ2)}

Table~\ref{tab:rq2} summarizes the performance of our proposed R-SFT approach and the subsequent merging step across various knowledge editing datasets in Knowedit. We observe that R-SFT consistently achieves near 100\% accuracy on the training samples and maintains approximately 60\% portability to reason with new knowledge, significantly outperforming conventional fine-tuning methods.

After model merging, the edited model consistently experiences a modest reduction (around 5\%) in editing accuracy, but this is acceptable given the restoration of the model's general capabilities. The complete result is provided in the Appendix~\ref{app:rq2}.

\begin{figure}[t]
    \centering
    \includegraphics[width=\linewidth]{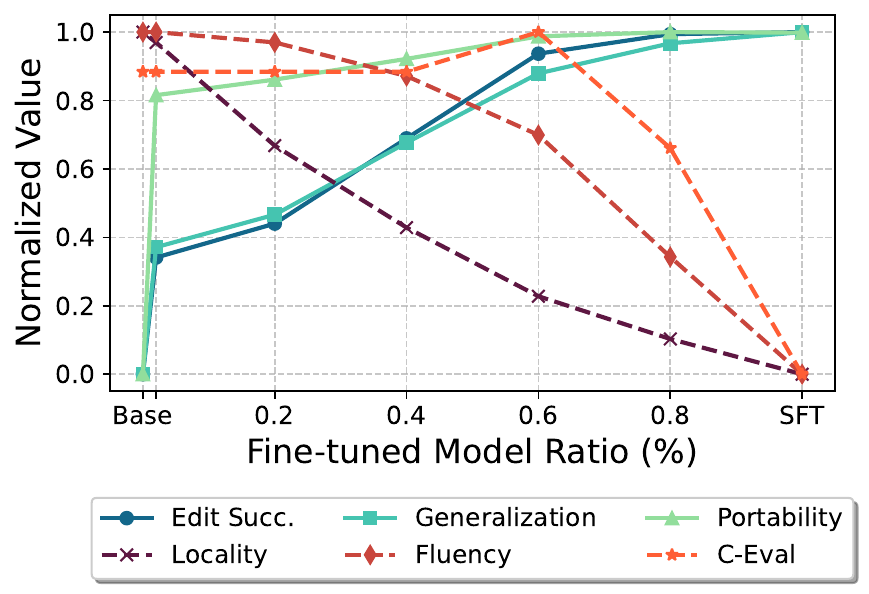}
    \caption{Metrics across different scaling ratios, illustrating the trade-off between edited and general knowledge.}
    \label{fig:scaling}
\end{figure}

\begin{figure}[t]
    \centering
    \includegraphics[width=\linewidth]{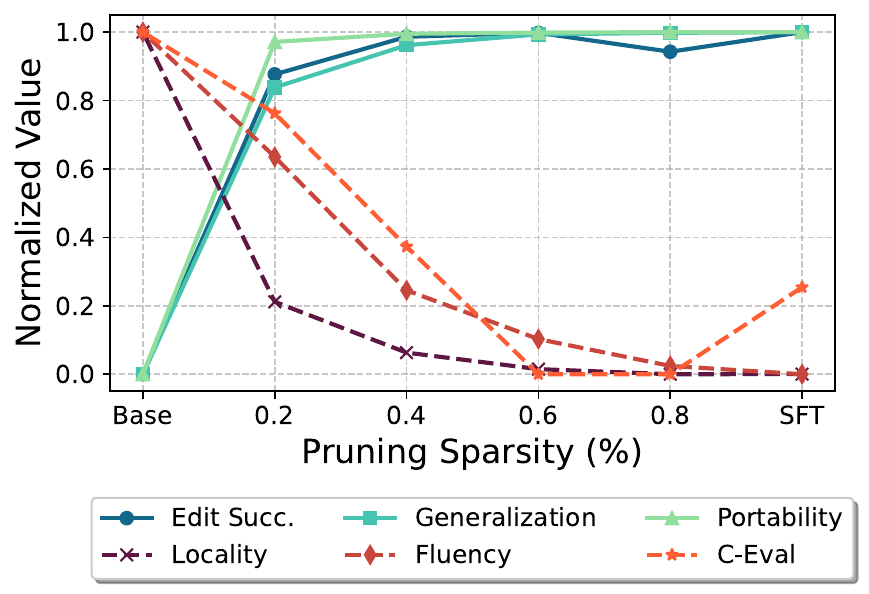}
    \caption{Metrics across different pruning sparseness, balancing edited and general knowledge.}
    \label{fig:pruning}
\end{figure}

\begin{table*}[t]
\footnotesize
\centering
\caption{Ablation study of the framework on editing performance (including success rate, generalization, portability, locality, and fluency) and general capabilities based on C-Eval (Acc.), CoQA (F1), and LogiQA (Acc.).}
\label{tab:ablation}
\resizebox{\textwidth}{!}{  
\begin{tabular}{@{}llcccccccc@{}}
\toprule
\textbf{Stage} & \textbf{Methods} & \textbf{Succ.} & \textbf{Gen.} & \textbf{Port.} & \textbf{Loc.} & \textbf{Flu.} & \textbf{C-Eval} & \textbf{CoQA} & \textbf{LogiQA} \\ \midrule
Base &  & - & - & - & - & - & 79.57 & 72.60 & 37.94 \\ \midrule
\multirow{3}{*}{R-SFT} & w/o Sample Steps & \textbf{99.82} & 93.85 & \textbf{47.32} & \textbf{35.03} & 466.00 & \textbf{44.28} & \textbf{63.57} & 24.73 \\
 & w/o Early Stop & \textbf{99.82} & \textbf{93.95} & 41.10 & 31.51 & \textbf{534.19} & 40.04 & 53.11 & 23.81 \\
 & Complete & 99.43 & 93.70 & 45.93 & 33.96 & 401.44 & 41.60 & 58.84 & \textbf{26.57} \\ \midrule
\multirow{3}{*}{Merging} & w/o Scaling & 98.25 & 92.36 & 45.14 & 33.96 & 411.70 & 58.47 & 62.00 & 32.41 \\
 & w/o Pruning & 96.97 & 92.07 & 42.76 & 29.69 & 418.32 & 52.75 & 74.65 & 29.80 \\
 & Complete & 96.95 & 91.58 & 39.63 & 26.42 & 420.49 & 68.42 & 78.07 & 34.25 \\ \bottomrule
\end{tabular}
}
\end{table*}

\subsection{Parameter Analysis (RQ3)}

\paragraph{R-SFT.} As shown in Tables~\ref{tab:early}, stopping training early (lower thresholds) improves generalization by preventing overfitting. A moderate threshold of 0.1 strikes the optimal balance between gaining knowledge and preventing overfitting. The results in Tables~\ref{tab:epoch-step} confirm that fewer steps per sample yield better performance. However, this approach requires absolute $E\times N\times K$ update steps, resulting in lower computational efficiency. Finally, five epochs with six steps per sample provide an optimal compromise. Appendix~\ref{app:pa-rsft} shows complete results for all hyperparameters.

\paragraph{Model Merging.} Figure~\ref{fig:scaling} and Figure~\ref{fig:pruning} demonstrate that scaling has a more immediate and pronounced impact on model performance, with an optimal setting typically around 0.8 to balance knowledge updates and generalization. In contrast, pruning exhibits a more subtle influence, and a sparsity ratio of 0.2 is generally preferred to minimize interference while preserving core capabilities.

\subsection{Ablation Study (RQ4)}

We conduct an ablation study to evaluate the individual contributions of each proposed component, as presented in Table~\ref{tab:ablation}. Results show that removing the sample-wise consecutive update (``w/o Sample Steps'') does not significantly harm editing performance, suggesting that our iterative update strategy does not negatively impact model quality while considerably enhancing efficiency. In contrast, removing early stopping (``w/o Early Stop'') significantly degrades the model's general capabilities, confirming its essential role in preventing overfitting. In the model merging stage, omitting either scaling (``w/o Scaling'') or pruning (``w/o Pruning'') leads to decreased restoration of general capabilities, highlighting the importance of these techniques in effectively balancing knowledge editing and general model performance.

\section{Related Works}
\label{related-works}

\subsection{Knowledge Editing}

Knowledge editing aims to efficiently update or modify the internal knowledge of machine learning models to adapt to rapidly changing real-world information~\cite{page,pair}. This is particularly important for LLMs, whose training demands substantial computational resources and time, making frequent pretraining impractical~\cite{delora}. Early studies focused on knowledge tracing to analyze and locate factual information stored within models before attempting edits~\cite{trace,XES3G5M,bias-trace}. ROME~\cite{ROME} fisrt directly modified neurons associated with specific facts in feed-forward layers. While ROME models can edit certain facts accurately, many real-life situations involve dynamic information that require perpetual model updates~\cite{sequential-rec,embed}. This necessitates the development of editing techniques that support persistent change
Subsequent approaches, like MEMIT \cite{memit2022} and r-ROME \cite{rrome}, enhanced editing precision and stability during sequential updates.

Other methods utilized fine-tuning on specialized datasets~\cite{delora}, effectively injecting knowledge but risking general capability degradation due to overfitting. Meta-learning approaches (e.g., MEND \cite{MEND}, InstructEdit \cite{instructedit2021}) and memory-based methods (e.g., SERAC \cite{SERAC}, MELO \cite{melo2023}) achieved better generalization but introduced auxiliary networks or structured memories, significantly increasing model complexity and limiting practical deployment.

\subsection{Model Merging}

Model merging techniques combine parameters from multiple models or training checkpoints into a unified model. This technique is more efficient than using several LLMs simultaneously~\cite{agent,ensemble-survey}. Early methods primarily relied on simple weight averaging~\cite{model-soup}, but subsequent work introduced more sophisticated strategies. For instance, SLERP \cite{slerp2023} proposed spherical interpolation between model parameters to mitigate geometric distortion inherent in linear interpolation methods. Task Arithmetic \cite{taskarith2023}, and its extensions, such as TIES \cite{ties2023} and DARE \cite{dare2023}, computed and combined task vectors, effectively tackling inter-model interference via sparsification, sign-consensus algorithms, adaptive pruning, and parameter rescaling. More recently, WISE \cite{wang2024wise} applied sparsification methods to fine-tuning for knowledge editing, effectively balancing edited knowledge and pre-trained information, but also introduced increased structural complexity.

\section{Conclusion}
\label{sec:conclusion}

In this paper, we propose a two-stage framework for knowledge editing that integrates robust supervised fine-tuning (R-SFT) with model merging. 
Specifically, R-SFT first leverages sample-wise iterative updates and an early-stopping mechanism to precisely inject new knowledge with enhanced generalization. 
Subsequently, the model merging technique serves to further mitigate the harm of fine-tuning by merging the pre-trained model with the R-SFT model, thus negating the necessity for architectural changes. 
Experimental results show that our method significantly outperforms existing approaches in sequential editing scenarios while maintaining general capabilities.

\section{Limitations}
\label{sec:limitations}

Although our model merging approach demonstrates significant effectiveness in knowledge editing, we acknowledge certain limitations in knowledge generalization capabilities. Our current framework, while successful at direct knowledge updates, shows reduced performance when transferring edited knowledge to substantially different phrasings or when applying reasoning based on newly acquired information. The generalization metrics indicate room for improvement in how edited knowledge is applied across varied contexts.
Future research should focus on developing more sophisticated knowledge insertion methods that enhance the transferability of edited information.

\section*{Acknowledgments}

This research was partially supported by Research Impact Fund (No.R1015-23), Collaborative Research Fund (No.C1043-24GF) and Tencent (CCF-Tencent Open Fund, Tencent Rhino-Bird Focused Research Program).

\bibliography{custom}

\appendix

\section{Detailed Experimental Settings}
\label{app:exp-settings}

\subsection{Datasets}

KnowEdit~\cite{knowedit} contains a total of six sub-datasets including Wiki$_{recent}$, ZsRE, WikiBio, WikiData$_{counterfact}$, Convsent and Sanitation.

For general ability evaluation, C-Eval~\cite{ceval} primarily assesses common knowledge, while other benchmarks are predominantly question-answering datasets designed to evaluate models' capabilities in extended conversations with longer textual contexts.

\subsection{Implementation Details}

During the training phase, we utilize a batch size of 1 to maximize the effective learning from each individual sample. Our R-SFT is configured with 5 epochs and 6 consecutive steps, employing a maximum learning rate of $5 \times 10^{-4}$.

\begin{table*}[t]
\footnotesize
\centering
\caption{Performance comparison of merging methods for sequential knowledge editing. The best values are highlighted in bold, while the second-best values are underlined.}
\label{app:tab-rq2}
\begin{tabular}{@{}lrcccccc@{}}
\toprule
\textbf{DataSet} & \textbf{Metric $\uparrow$} & \textbf{ROME} & \textbf{MEMIT} & \textbf{LoRA} & \textbf{SFT} & \textbf{R-SFT} & \textbf{Merged} \\ \midrule
\multirow{4}{*}{\textbf{WikiData$_{recent}$}} & Edit Succ. & 15.78 & 0.00 & 1.11 & 79.46 & \textbf{99.97} & {\ul 96.62} \\
 & Portability & 4.79 & 0.00 & 0.90 & 46.59 & {\ul 58.26} & \textbf{62.95} \\
 & Locality & 1.76 & 0.00 & 0.06 & 28.50 & {\ul 31.87} & \textbf{41.62} \\
 & Fluency & {\ul 529.98} & 478.64 & 505.02 & 428.95 & 461.51 & \textbf{592.02} \\ \midrule
\multirow{3}{*}{\textbf{WikiBio}} & Edit Succ. & 26.47 & 0.04 & 53.26 & 66.06 & \textbf{99.48} & {\ul 96.54} \\
 & Locality & 3.50 & 0.03 & {\ul 64.56} & 40.16 & 64.30 & \textbf{75.18} \\
 & Fluency & 608.15 & 502.35 & {\ul 627.18} & 626.60 & \textbf{628.77} & 626.71 \\ \midrule
\multirow{4}{*}{\textbf{WikiData$_{counter}$}} & Edit Succ. & 12.69 & 0.00 & 11.07 & 50.67 & \textbf{99.06} & {\ul 84.02} \\
 & Portability & 2.88 & 0.00 & 10.28 & 34.56 & \textbf{60.61} & {\ul 51.98} \\
 & Locality & 0.92 & 0.00 & 13.65 & 15.75 & {\ul 26.36} & \textbf{41.98} \\
 & Fluency & 553.18 & 314.91 & 489.65 & 479.81 & {\ul 601.02} & \textbf{614.64} \\ \bottomrule
\end{tabular}
\end{table*}

\subsection{Evaluation Metrics}
\label{ssec:app-metrcs}

For evaluating the editing performance of the merged models, we adopt four widely used metrics:
\begin{itemize}[leftmargin=*]
    \item \textbf{Edit Succ. (Succ.):} This metric quantifies whether the intended factual update is correctly reflected in the model’s output when given the edited query.
    \item \textbf{Generalization (Gen.):} This metric evaluates whether the model can correctly apply the updated factual knowledge when presented with semantically equivalent queries.
    \item \textbf{Portability (Port.):} This measures the ability of the edited model to generalize the new knowledge to alternative phrasings or reworded versions of the original query.
    \item \textbf{Locality (Loc.):} Locality evaluates whether the editing process is confined to the targeted knowledge, ensuring that the model's outputs for unrelated queries remain unchanged.
    \item \textbf{Fluency (Flu.):} This metric assesses the linguistic quality of the model's responses, verifying that the edited outputs are coherent and natural.
\end{itemize}

To comprehensively assess the general capabilities of the models after knowledge editing, we employ several established benchmarks with the following metrics:

\begin{itemize}[leftmargin=*]
    \item \textbf{Accuracy:} For classification tasks such as C-Eval and LogiQA, we utilize accuracy as the primary metric, which measures the percentage of correctly answered questions.
    \item \textbf{Exact Match (EM):} For extractive question answering tasks including CoQA, DROP, and SQuAD 2.0, we report the Exact Match score, which requires the model's prediction to exactly match the ground truth answer:
    \begin{equation}
        \text{EM}(\mathbf{a}, \hat{\mathbf{a}}) = \mathbf{1}(\mathbf{a} = \hat{\mathbf{a}})
    \end{equation}
    where $\mathbf{a}$ is the ground truth answer, $\hat{\mathbf{a}}$ is the model's prediction, and $\mathbf{1}(\cdot)$ is the indicator function that returns 1 if the condition is true and 0 otherwise.
    
    \item \textbf{F1 Score (F1):} For the same question answering tasks, we also report the F1 score, which measures the overlap between the predicted and ground truth answers at the token level:
    \begin{equation}
        \text{F1} = \frac{2 \times \text{Precision} \times \text{Recall}}{\text{Precision} + \text{Recall}}
    \end{equation}
    where:
    \begin{equation}
        \text{Precision} = \frac{|\text{Tokens in } \hat{\mathbf{a}} \cap \text{Tokens in } \mathbf{a}|}{|\text{Tokens in } \hat{\mathbf{a}}|}
    \end{equation}
    \begin{equation}
        \text{Recall} = \frac{|\text{Tokens in } \hat{\mathbf{a}} \cap \text{Tokens in } \mathbf{a}|}{|\text{Tokens in } \mathbf{a}|}
    \end{equation}
\end{itemize}

\section{Knowledge Editing Performance (RQ2)}
\label{app:rq2}

Table \ref{app:tab-rq2} compares our approach against baseline knowledge editing methods. Our R-SFT consistently achieves the highest editing success rates while maintaining strong portability. The merged model, while showing slightly lower editing success than R-SFT, demonstrates superior locality and fluency, effectively balancing edit fidelity with preservation of general capabilities. Parameter-efficient methods (ROME, MEMIT, LoRA) that perform well in single-fact editing struggle significantly in sequential editing scenarios, highlighting our framework's advantage in practical applications requiring both accurate knowledge editing and maintained model quality.

\section{Parameter Analysis of R-SFT (RQ3)}
\label{app:pa-rsft}

\begin{table}[t]
\centering
\caption{Effect of edited layer selection on knowledge editing performance.}
\label{tab:layer-selection}
\resizebox{\linewidth}{!}{
\begin{tabular}{@{}lccccc@{}}
\toprule
\textbf{Layer} & \textbf{Succ.} & \textbf{Gen.} & \textbf{Port.} & \textbf{Loc.} & \textbf{Flu.} \\ \midrule
5 & 75.74 & 73.28 & 39.86 & 27.84 & 435.20 \\
6 & \textbf{85.49} & \textbf{83.38} & 41.85 & 31.97 & 431.43 \\
7 & 85.31 & 81.81 & \textbf{44.08} & \textbf{34.61} & 434.13 \\
13 & 74.58 & 68.61 & 38.07 & 33.87 & 492.87 \\
20 & 70.03 & 62.37 & 26.43 & 21.55 & \textbf{497.90} \\
27 & 56.97 & 52.44 & 18.39 & 8.08 & 385.88 \\ \bottomrule
\end{tabular}
}
\end{table}

\begin{table}[t]
\centering
\caption{Effect of maximum training steps per sample on editing performance.}
\label{tab:training-steps}
\resizebox{\linewidth}{!}{
\begin{tabular}{@{}lccccc@{}}
\toprule
\textbf{Steps} & \textbf{Succ.} & \textbf{Gen.} & \textbf{Port.} & \textbf{Loc.} & \textbf{Flu.} \\ \midrule
30 & 75.74 & 73.28 & 39.86 & 27.84 & 435.20 \\
60 & 75.74 & 73.28 & 39.86 & 27.84 & 435.20 \\
90 & 75.74 & 73.28 & 39.86 & 27.84 & 435.20 \\ \bottomrule
\end{tabular}
}
\end{table}

\begin{table}[ht]
\centering
\caption{Effect of the number of edited layers on editing performance.}
\label{tab:num-layers}
\resizebox{\linewidth}{!}{
\begin{tabular}{@{}lccccc@{}}
\toprule
\textbf{Layers} & \textbf{Succ.} & \textbf{Gen.} & \textbf{Port.} & \textbf{Loc.} & \textbf{Flu.} \\ \midrule
Layer 5 & \textbf{75.74} & \textbf{73.28} & \textbf{39.86} & \textbf{27.84} & \textbf{435.20} \\
Layers 4,5,6 & 66.96 & 62.95 & 28.36 & 16.64 & 409.75 \\
All Layers & 12.93 & 12.62 & 4.27 & 1.85 & 380.84 \\ \bottomrule
\end{tabular}
}
\end{table}

\begin{table}[t]
\centering
\caption{Effect of learning rate (LR.) on editing performance.}
\label{tab:learning-rate}
\begin{tabular}{@{}lccccc@{}}
\toprule
\textbf{LR.} & \textbf{Succ.} & \textbf{Gen.} & \textbf{Port.} & \textbf{Loc.} & \textbf{Flu.} \\ \midrule
5e-4 & \textbf{75.74} & \textbf{73.28} & 39.86 & 27.84 & 435.20 \\
1e-4 & 67.68 & 61.75 & \textbf{48.33} & 41.55 & 516.84 \\
5e-5 & 63.12 & 54.90 & 45.97 & \textbf{44.11} & \textbf{556.84} \\ \bottomrule
\end{tabular}
\end{table}

\paragraph{Edited Layer Selection}

Table~\ref{tab:layer-selection} presents the performance when editing different layers of the LLM. Layers 6 and 7 consistently outperform other layers across most metrics, with Layer 6 achieving the highest edit success (85.49\%) and generalization (83.38\%). This result confirms findings from prior research that knowledge is more concentrated in the earlier layers of the LLM~\cite{ROME}.

\paragraph{Training Steps} Table~\ref{tab:training-steps} examines how many total steps are typically required to update each sample when early stopping is enabled. With early stopping enabled (loss threshold = 0.01), we observe that performance metrics remain identical across different maximum step settings. This indicates that typically within 30 steps the loss of one sample will converge.

\paragraph{Number of Edited Layers} Table~\ref{tab:num-layers} investigates the impact of simultaneously editing multiple layers versus focusing on a single layer. Contrary to intuition, editing a single layer (Layer 5) yields substantially better results than editing multiple layers. Editing all layers leads to catastrophic performance degradation across all metrics. This suggests that targeted, minimal interventions are more effective for knowledge editing than widespread parameter modifications.

\paragraph{Learning Rate} Table~\ref{tab:learning-rate} examines how different learning rates affect the editing process. Our analysis reveals an interesting trade-off: higher learning rates (5e-4) improve edit success and generalization but reduce portability, locality, and fluency. Conversely, lower learning rates (5e-5) significantly enhance fluency and locality at the expense of edit success and generalization. This suggests that the optimal learning rate depends on which metrics are prioritized for a specific application.



\end{document}